\documentclass[11pt,a4paper]{article}
\usepackage[hyperref]{acl2021}
\usepackage{times}
\usepackage{latexsym}

\usepackage{microtype}
\usepackage{graphicx}
\usepackage{booktabs}  
\usepackage{multirow}
\usepackage{subfigure}
\usepackage{verbatim}
\usepackage{CJKutf8} 
\usepackage{flushend}
\usepackage{makecell}
\aclfinalcopy 

\title{One Teacher is Enough? \\ Pre-trained Language Model Distillation from Multiple Teachers}
\author{Chuhan Wu$^\dagger$~~~~Fangzhao Wu$^\ddagger$~~~~Yongfeng Huang$^\dagger$\\
    $^\dagger$Department of Electronic Engineering \& BNRist, Tsinghua University, Beijing 100084, China  \\
     $^\ddagger$Microsoft Research Asia, Beijing 100080, China\\
  \tt{\{wuchuhan15, wufangzhao\}@gmail.com, yfhuang@tsinghua.edu.cn}
  }
\date{}

\begin{document}
\maketitle
\begin{abstract}

Pre-trained language models (PLMs) achieve great success in NLP.
However, their huge model sizes hinder their applications in many practical systems.
Knowledge distillation is a popular technique to compress PLMs, which learns a small student model from a large teacher PLM.
However, the knowledge learned from a single teacher may be limited and even biased, resulting in low-quality student model.
In this paper, we propose a multi-teacher knowledge distillation framework named MT-BERT for pre-trained language model compression, which can train high-quality student model from multiple teacher PLMs.
In MT-BERT we design a multi-teacher co-finetuning method to jointly finetune multiple teacher PLMs in downstream tasks with shared pooling and prediction layers to align their output space for better collaborative teaching.
In addition, we propose a multi-teacher hidden loss and a multi-teacher distillation loss to transfer the useful knowledge in both hidden states and soft labels from multiple teacher PLMs to the student model.
Experiments on three benchmark datasets validate the effectiveness of MT-BERT in compressing PLMs.

\end{abstract}

\section{Introduction}

Pre-trained language models (PLMs) such as BERT and RoBERTa have achieved notable success in various NLP tasks~\cite{devlin2019bert,yang2019xlnet,liu2019roberta}.
However, many PLMs have a huge model size and computational complexity, making it difficult to deploy them to low-latency and high-concurrence online systems or devices with limited computational resources~\cite{jiao2020tinybert,wu2021newsbert}.

Knowledge distillation is a widely used technique for compressing large-scale pre-trained language models~\cite{sun2019patient,wang2020minilm}.
For example,~\citet{sanh2019distilbert} proposed DistilBERT to compress BERT by transferring knowledge from the soft labels predicted by the teacher model to student model with a distillation loss.
\citet{jiao2020tinybert} proposed TinyBERT, which aligns the hidden states and the attention heatmaps between student and teacher models.
These methods usually learn the student model from a single teacher model~\cite{gou2020knowledge}.
However, the knowledge and supervision provided by a single teacher model may be insufficient to learn an accurate student model, and the student model may also inherit the bias in the teacher model~\cite{bhardwaj2020bias}.
Fortunately, many different large PLMs such as BERT~\cite{devlin2019bert}, RoBERTa~\cite{liu2019roberta} and UniLM~\cite{dong2019unilm} are off-the-shelf.
These PLMs may encode complementary knowledge because they usually have different configurations and are trained on different corpus with different self-supervision tasks~\cite{qiu2020pre}.
Thus, incorporating multiple pre-trained language models into knowledge distillation has the potential to learn better student models.

In this paper, we present a multi-teacher knowledge distillation method named MT-BERT for pre-trained language model compression.\footnote{We focus on task-specific knowledge distillation.}
In MT-BERT, we propose a multi-teacher co-finetuning framework to jointly finetune multiple teacher models with a shared pooling and prediction module to align their output hidden states for better collaborative student teaching. 
In addition, we propose a multi-teacher hidden loss and a multi-teacher distillation loss to transfer the useful knowledge in both hidden states and soft labels from multiple teacher models to student model.
Experiments on three benchmark datasets show MT-BERT can effectively improve the quality of student models for PLM compression and outperform many single-teacher knowledge distillation methods.

\section{MT-BERT}\label{sec:Model}

Next, we introduce the details of our multi-teacher knowledge distillation method MT-BERT for pre-trained language model compression.\footnote{Codes available at https://github.com/wuch15/MT-BERT}
We first introduce the multi-teacher co-finetuning framework to jointly finetune multiple teacher models in downstream tasks, and then introduce the multi-teacher distillation framework to collaboratively teach the student with multiple teachers.

\subsection{Multi-Teacher Co-Finetuning}

\begin{figure}[!t]
  \centering 
      \includegraphics[width=0.98\linewidth]{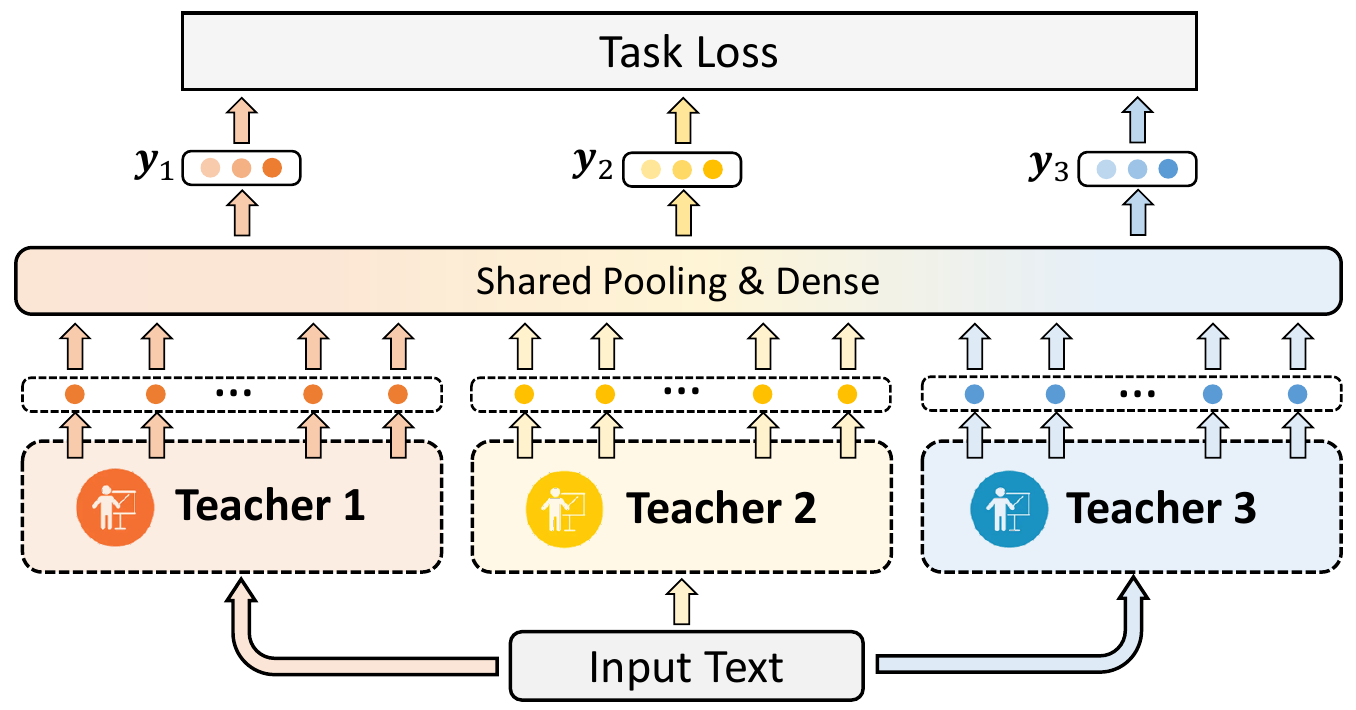}
  \caption{The multi-teacher co-finetuning framework.}\label{fig.finetune}
\end{figure}

Researchers have found that distilling the knowledge in the hidden states of a teacher model is important for effective student teaching~\cite{sun2019patient,jiao2020tinybert}.
However, since different teacher PLMs are separately pre-trained with different settings, finetuning them independently may lead to some inconsistency in their feature space, which is not optimal for transferring knowledge in the hidden states of multiple teachers.
Thus, we design a multi-teacher co-finetuning framework to obtain some uniformity among the hidden states output by the last layer of different teacher models for better collaborative student teaching, as shown in Fig.~\ref{fig.finetune}.
Assume there are $N$ teacher models, and  denote the hidden states output by the top layer of the $i$-th teacher as $\mathbf{H}^i$.
We use a shared pooling\footnote{In MT-BERT we use attentive pooling because it performs better than average pooling and ``[CLS]'' token embedding.
} layer to summarize each hidden matrix $\mathbf{H}^i$ into a unified text embedding, and then use a shared dense layer to convert it into a soft probability vector $\mathbf{y}_i$.
Finally, we jointly optimize the summation of the task-specific losses of all teacher models, i.e., $\sum_{i=1}^N \rm{CE}(\mathbf{y},\mathbf{y}_i)$, where $\rm{CE}(\cdot, \cdot)$ stands for the cross-entropy loss and $\mathbf{y}$ is the ground-truth label.
Since the pooling and prediction layers are shared among different teachers, the feature space of the output hidden states from different teacher PLMs can be aligned, which can help them collaborate better for student teaching.

\subsection{Multi-Teacher Knowledge Distillation}

\begin{figure*}[!t]
  \centering 
      \includegraphics[width=0.99\linewidth]{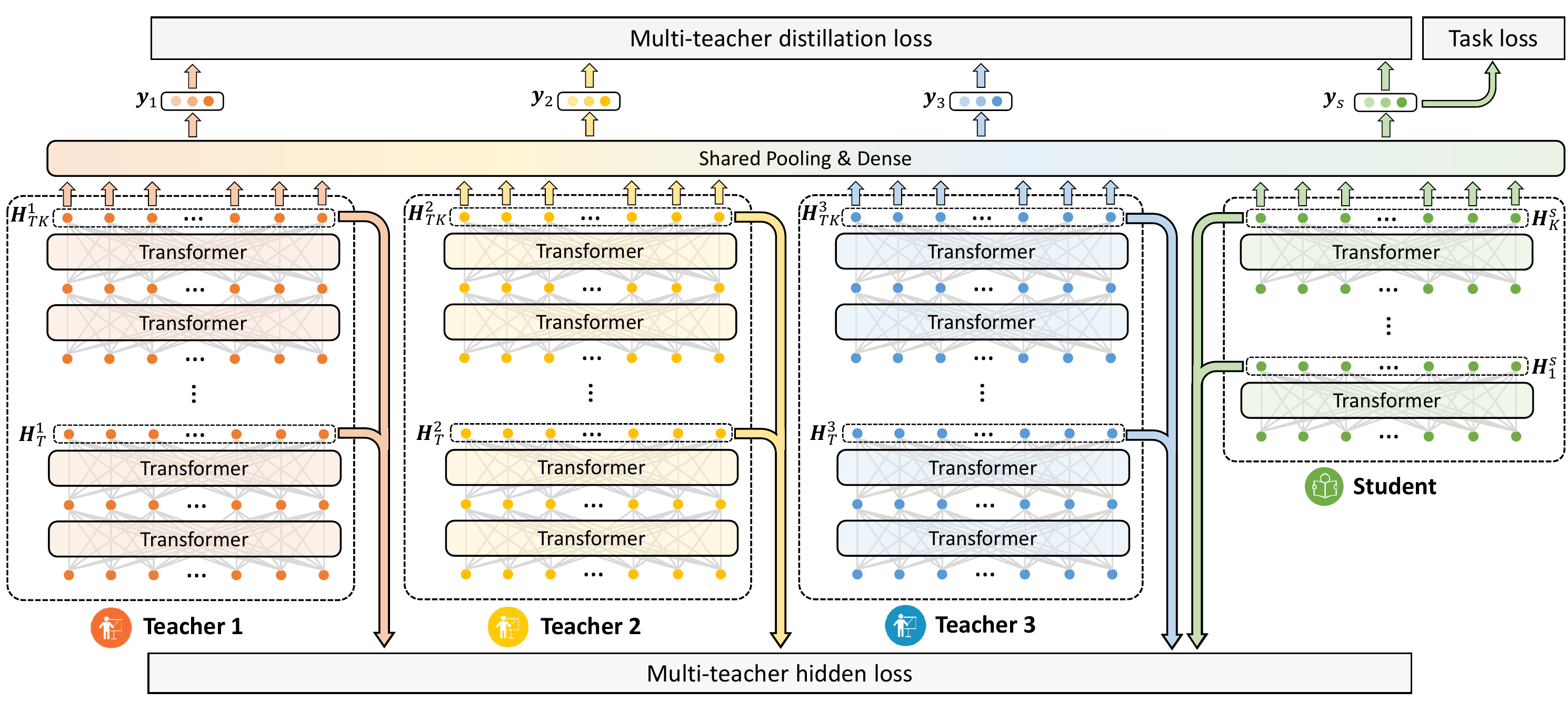}
  \caption{The multi-teacher knowledge distillation framework in MT-BERT.}\label{fig.mtbert}
\end{figure*}

Next, we introduce our proposed multi-teacher knowledge distillation framework, which is shown in Fig.~\ref{fig.mtbert}.
Two loss functions are used for knowledge distillation, i.e., a multi-teacher hidden loss and a multi-teacher distillation loss.

The multi-teacher hidden loss aims to transfer knowledge in the hidden states of multiple teachers. 
Assume there are $N$ teacher PLMs, and each of them has $T\times K$ Transformer layers.
They collaboratively teach a student model with $K$ layers, and each layer in the student model corresponds to $T$ layers in teacher PLMs.\footnote{Here we assume that all teacher models have the same number of layers. We will explore to generalize MT-BERT to scenarios where teacher models have different architectures in our future work.}
Denote the hidden states output by the $j$-th layer of the student model as $\mathbf{H}^s_j$, and the corresponding hidden states output by the $(T\times j)$-th layer of the $i$-th teacher model as $\mathbf{H}^i_{Tj}$.
Following~\cite{sun2019patient}, we apply the mean squared error (MSE) to the hidden states of corresponding layers in the student and teacher models to encourage the student model to have similar functions with teacher models.
The multi-teacher hidden loss $\mathcal{L}_{MT-Hid}$ is formulated as follows:
\begin{equation}
    \mathcal{L}_{MT-Hid}=\sum_{i=1}^N\sum_{j=1}^T \rm{MSE}(\mathbf{H}^s_j, \mathbf{W}_{ij}\mathbf{H}^i_{Tj}),
\end{equation}
where $\mathbf{W}_{ij}$ is a learnable  transformation matrix.

The multi-teacher distillation loss aims to transfer the knowledge in the soft labels output by multiple teachers to student. 
The predictions of different teachers on the same sample may have different correctness and confidence.
Thus, it may be suboptimal to simply ensemble~\cite{fukuda2017efficient,liu2020adaptive} or choose~\cite{yuan2020reinforced} soft labels without the help of task labels.
Since in task-specific knowledge distillation the labels of training samples are available, we propose a distillation loss weighting method to assign different weights to different samples.
The weights are based on the loss inferred from the predictions of corresponding teacher against the gold labels.
More specifically, the multi-teacher distillation loss $\mathcal{L}_{MT-Dis}$ is formulated as follows:
\begin{equation}\label{mtdis}
    \mathcal{L}_{MT-Dis}=\sum_{i=1}^N \frac{\rm{CE}(\mathbf{y}_i/t, \mathbf{y}_s/t)}{1+\rm{CE}(\mathbf{y},\mathbf{y}_i)},
\end{equation}
where $t$ is the temperature coefficient.
In this way, if a teacher's prediction on a certain sample is more close to the ground-truth label, its corresponding distillation loss will gain higher weight.

Following~\cite{tang2019distilling,lu2020twinbert}, we also incorporate gold labels to compute the task-specific loss $\mathcal{L}_{Task}$ based on the predictions of the student model, i.e., $\mathcal{L}_{Task}=\rm{CE}(\mathbf{y},\mathbf{y}_s)$.
The final loss function $\mathcal{L}$ for learning the student model is a summation of the  multi-teacher hidden loss, multi-teacher distillation loss and the task-specific loss, which is formulated as follows:
\begin{equation}
    \mathcal{L}=\mathcal{L}_{MT-Hid}+\mathcal{L}_{MT-Dis}+\mathcal{L}_{Task}.
\end{equation}

\section{Experiments}\label{sec:Experiments}

\subsection{Datasets and Experimental Settings}

We conduct experiments on three benchmark datasets with different sizes.
The first one is SST-2~\cite{socher2013recursive}, which is a benchmark for text sentiment classification.
The second one is RTE~\cite{bentivogli2009fifth}, which is a widely used dataset for natural language inference.
The third one is the MIND dataset~\cite{wu2020mind}, which is a large-scale public English news dataset.\footnote{https://msnews.github.io/}
We perform the news topic classification task on this dataset.
The detailed statistics of the three datasets are shown in Table~\ref{dataset}.

\begin{table}[h]
\centering
\resizebox{0.445\textwidth}{!}{ 
\begin{tabular}{lcccc}
 \Xhline{1.5pt}
\textbf{Dataset} & \textbf{\#Train} &  \textbf{\#Dev} &  \textbf{\#Test}  &  \textbf{\#Class} \\ \hline
SST-2   & 67k                         & 872                       & 1.8k                       & 2                           \\
RTE     & 2.5k                        & 276                       & 3.0k                       & 2                           \\
MIND    & 102k                        & 2.6k                      & 26k                        & 18                          \\  
 \Xhline{1.5pt}
\end{tabular}
}
\caption{The statistics of the three datasets.}\label{dataset}
\end{table}

In our experiments, we use the pre-trained 12-layer BERT, RoBERTa and UniLM~\cite{bao2020unilmv2}\footnote{We used the UniLMv2 version.} models as the teachers to distill a 6-layer and a 4-layer student models respectively.
We use the token embeddings and the first 4 or 6 Transformer layers of UniLM to initialize the parameters of the student model.
The pooling layer is implemented by an attention network~\cite{yang2016hierarchical,wu2020attentive}.
The temperature coefficient $t$ is set to 1.
The attention query dimension in the attentive pooling layer is 200.
The optimizer we use is Adam~\cite{kingma2014adam}.
The teacher model learning rate  is 2e-6 while the
student model learning rate is 5e-6.
The batch size is 64.
Following~\cite{jiao2020tinybert}, we report the accuracy score on the SST-2 and RTE datasets.
In addition, since the news topics in the MIND dataset are highly imbalanced, following~\cite{wu2020improving} we report both accuracy and macro-F1 scores.
Each experiment is independently repeated 5 times and  the average scores are reported.

\begin{table}[!t]
\resizebox{0.48\textwidth}{!}{
\begin{tabular}{lccccc}
\Xhline{1.5pt}
\textbf{Methods}     & \begin{tabular}[c]{@{}c@{}}\textbf{SST-2}\\ (Acc.)\end{tabular} & \begin{tabular}[c]{@{}c@{}}\textbf{RTE}\\ (Acc.)\end{tabular} & \multicolumn{2}{c}{\begin{tabular}[c]{@{}c@{}}\textbf{MIND}\\ (Acc./Macro-F)\end{tabular}} & \textbf{\#Param} \\ \hline
BERT$_{12}$      & 92.8                                                   & 68.6                                                 & 73.6                                    & 51.3                                    & 109M    \\
RoBERTa$_{12}$   & 94.8                                                   & 78.7                                                 & 73.9                                    & 51.5                                    & 109M    \\
UniLM$_{12}$     & \textbf{95.1}                                                   & \textbf{81.3}                                                 & \textbf{74.6}                                    & \textbf{51.9}                                    & 109M    \\ \hline
DistilBERT$_{6}$ & 92.5                                                   & 58.4                                                 & 72.5                                    & 50.4                                    & 67.0M   \\
DistilBERT$_4$ & 91.4                                                   & 54.1                                                 & 72.1                                    & 50.2                                    & 52.2M   \\
BERT-PKD$_{6}$   & 92.0                                                   & 65.5                                                 & 72.7                                    & 50.6                                    & 67.0M   \\
BERT-PKD$_4$   & 89.4                                                   & 62.3                                                 & 72.4                                    & 50.3                                    & 52.2M   \\
TinyBERT$_{6}$   & 93.1                                                   & 70.0                                                 & 73.4                                    & 50.8                                    & 67.0M   \\
TinyBERT$_4$   & 92.6                                                   & 66.6                                                 & 73.0                                    & 50.4                                    & 14.5M   \\ \hline
MT-BERT$_{6}$     & \textbf{94.6}                                                   & \textbf{75.7}                                                 & \textbf{74.0}                                    & \textbf{51.5}                                    & 67.0M   \\
MT-BERT$_4$     & 93.9                                                   & 73.8                                                 & 73.8                                    & 51.2                                    & 52.2M   \\ \Xhline{1.5pt}
\end{tabular}
}
\caption{Results and parameters of different methods.} \label{table.performance} 
\end{table}

\begin{figure}[!t]
  \centering
  \subfigure[SST-2 and RTE datasets.]{
    \includegraphics[width=0.445\textwidth]{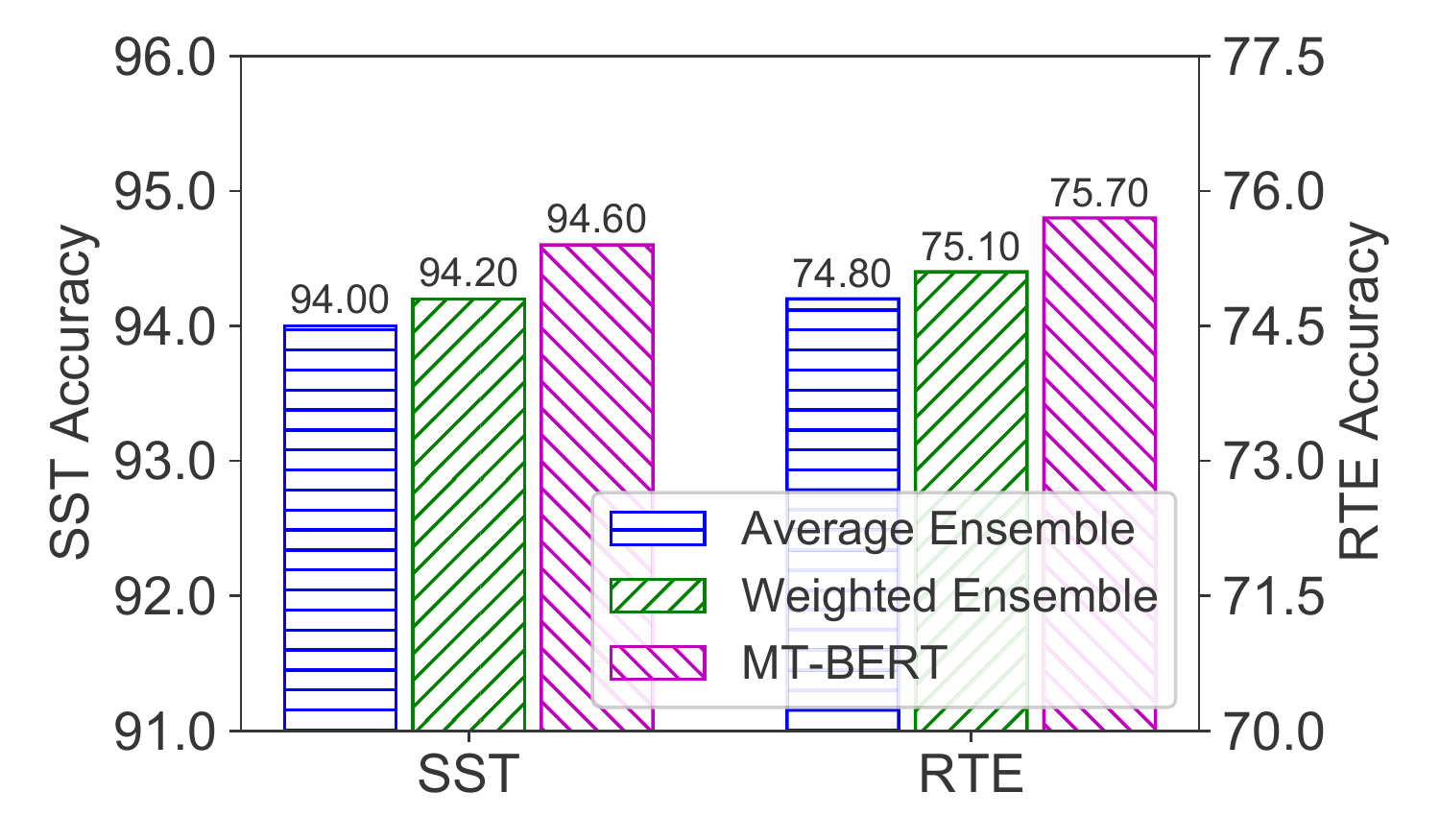}
  \label{fig.en1}
  }
   \subfigure[MIND dataset.]{
      \includegraphics[width=0.445\textwidth]{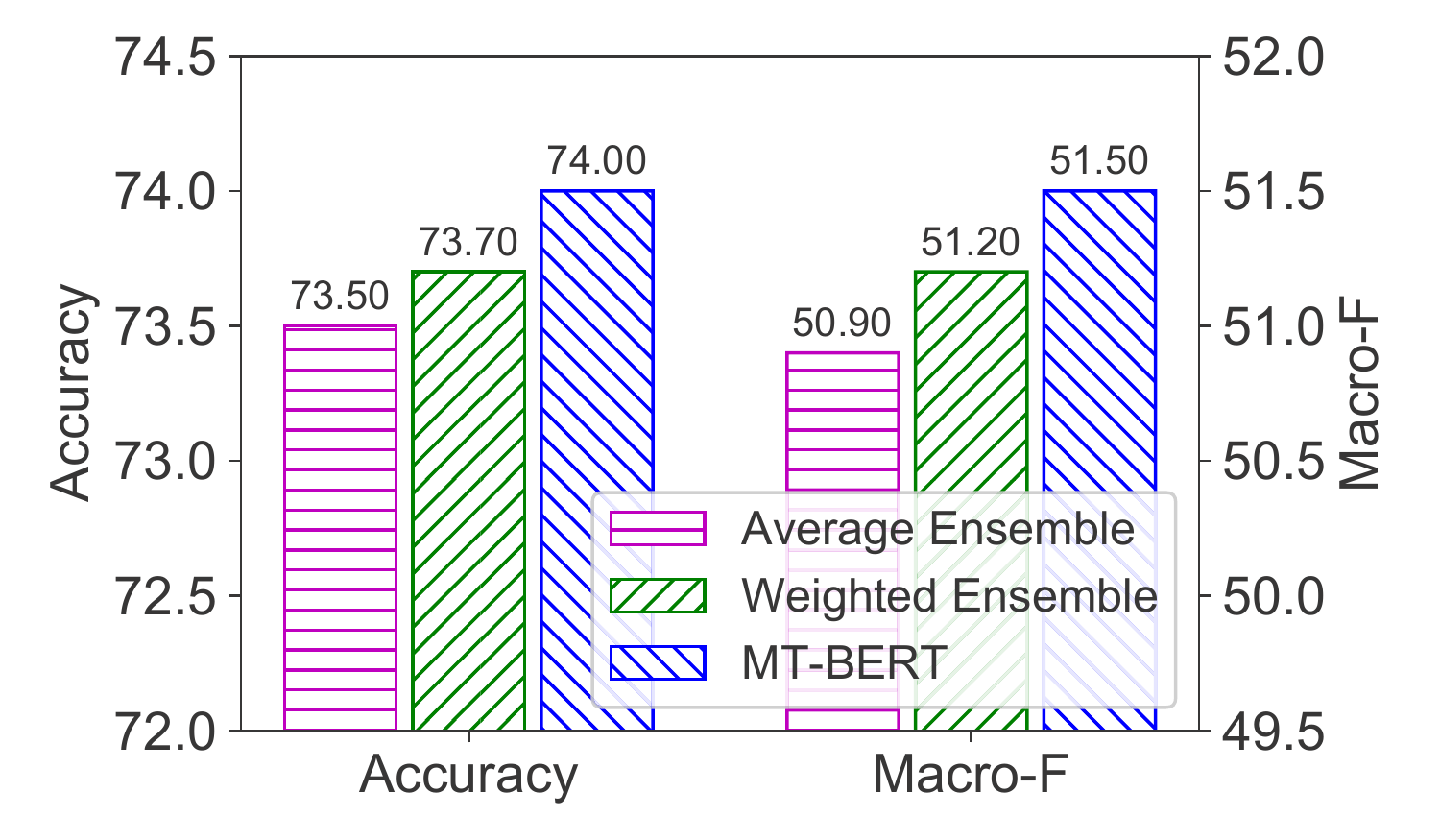}
  \label{fig.en2}
  }
  \caption{Comparison of MT-BERT and ensemble-based multi-teacher distillation methods.}\label{fig.en}
\end{figure}

\subsection{Performance Evaluation}

We compare the performance of MT-BERT with two groups of baselines.
The first group includes the 12-layer version of the teacher models, i.e., BERT~\cite{devlin2019bert}, RoBERTa~\cite{liu2019roberta} and UniLM~\cite{bao2020unilmv2}.
The second group includes the 6-layer and 4-layer student models distilled by  DistilBERT~\cite{sanh2019distilbert}, BERT-PKD~\cite{sun2019patient} and TinyBERT~\cite{jiao2020tinybert}, respectively.
The results of different methods are summarized in Table~\ref{table.performance}.\footnote{We take the original reported results of baseline methods on the SST-2 and RTE datasets, and we run their codes to obtain their results on the \textit{MIND} dataset.}
Referring to this table, we find MT-BERT can consistently outperform all the single-teacher knowledge distillation methods compared here.
This is because the knowledge provided by a single teacher model may be insufficient, and incorporating the complementary knowledge encoded in multiple teacher models can help learn better student model.
In addition, compared with the teacher models, MT-BERT has much fewer parameters and its performance is comparable or even better than these teacher models.
It shows that MT-BERT can effectively inherit the knowledge of multiple teacher models even if the model size is significantly compressed.

We also compare MT-BERT with several multi-teacher knowledge distillation methods proposed in the computer vision field that ensemble the outputs of different teachers for student teaching~\cite{you2017learning,liu2020adaptive}.
The results are shown in Fig.~\ref{fig.en}.
We find our MT-BERT performs better than these ensemble-based multi-teacher knowledge distillation methods.
This is because these methods do not consider the correctness of the teacher model predictions on a specific sample and cannot transfer useful knowledge encoded in the intermediate layers, which may not be optimal for collaborative knowledge distillation from multiple teachers.

\begin{table}[t]
\resizebox{0.48\textwidth}{!}{
\begin{tabular}{lcccc}
\Xhline{1.5pt}
\textbf{Teachers}      & \begin{tabular}[c]{@{}c@{}}\textbf{SST-2}\\ (Acc.)\end{tabular} & \begin{tabular}[c]{@{}c@{}}\textbf{RTE}\\ (Acc.)\end{tabular} & \multicolumn{2}{c}{\begin{tabular}[c]{@{}c@{}}\textbf{MIND}\\ (Acc./Macro-F)\end{tabular}} \\ \hline
BERT          & 92.1                                                   & 65.8                                                 & 72.8                                    & 50.6                                    \\
RoBERTa       & 92.9                                                   & 68.9                                                 & 73.0                                    & 50.7                                    \\
UniLM         & 93.3                                                   & 70.6                                                 & 73.4                                    & 50.9                                    \\ \hline
BERT+RoBERTa  & 93.6                                                   & 71.2                                                 & 73.3                                    & 50.9                                    \\
BERT+UniLM    & 93.9                                                   & 73.7                                                 & 73.6                                    & 51.1                                    \\
RoBERTa+UniLM & 94.3                                                   & 74.9                                                 & 73.7                                    & 51.3                                    \\ \hline
All           & 94.6                                                   & 75.7                                                 & 74.0                                    & 51.5                                    \\ \Xhline{1.5pt}
\end{tabular}
}
 \caption{Different combinations of teacher models.} \label{table.combine} 

\end{table}

\subsection{Effectiveness of Multiple Teachers}

Next, we study the effectiveness of using multiple teacher PLMs for knowledge distillation.
We compare the performance of the 6-layer student model distilled from different combinations of teacher models.
The results are summarized in Table~\ref{table.combine}.
It shows that using multiple teacher PLMs can achieve better performance than using a single one.
This is because different teacher models can encode complementary knowledge and combining them together can provide better supervision for student model.
In addition, combining all three teacher PLMs can further improve the performance of student model, which validates the effectiveness of MT-BERT in distilling knowledge from multiple teacher models.

\subsection{Ablation Study}

\begin{figure}[!t]
  \centering
  \subfigure[SST-2 and RTE datasets.]{
    \includegraphics[width=0.445\textwidth]{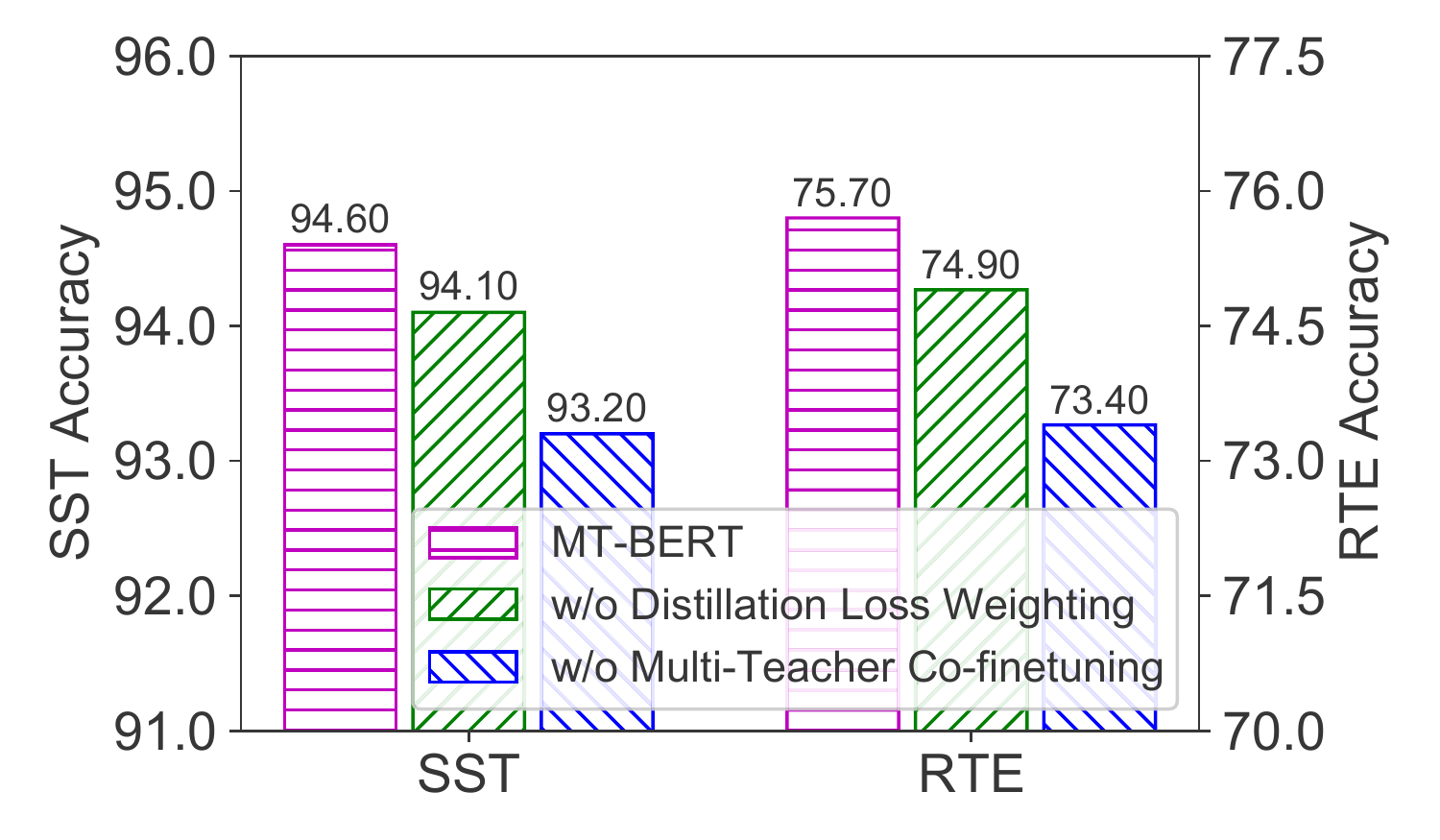}
  \label{fig.ab1}
  }
   \subfigure[MIND dataset.]{
      \includegraphics[width=0.445\textwidth]{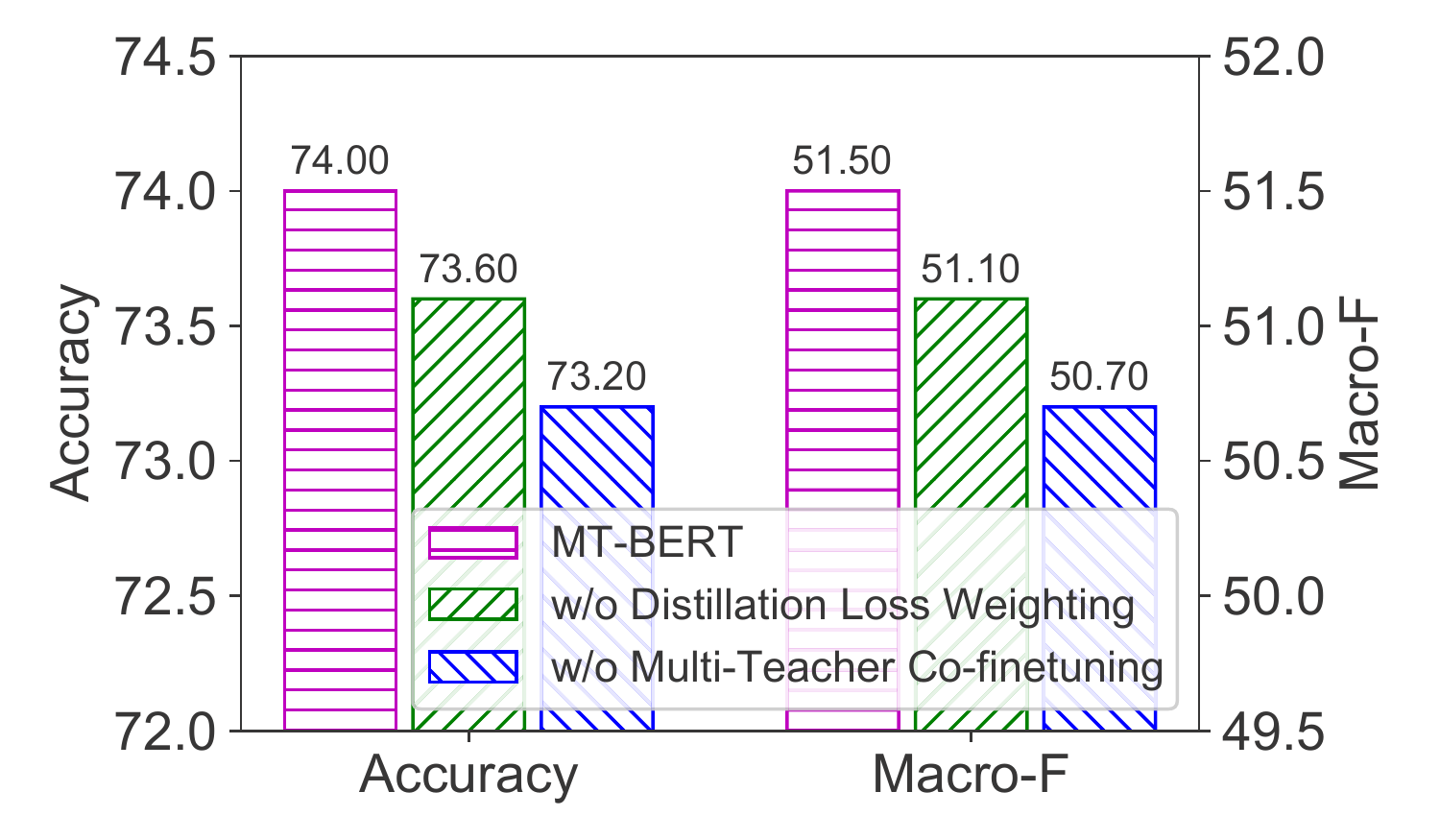}
  \label{fig.ab2}
  }
  \caption{Effectiveness of multi-teacher co-finetuning and distillation loss weighting.}\label{fig.ab}
\end{figure}

We study the effectiveness of the two important techniques in MT-BERT, i.e., the multi-teacher co-finetuning framework and the distillation loss weighting method.
We compare MT-BERT and its variants with one of these modules removed, as shown in Fig.~\ref{fig.ab}.
The student model has 6 layers.
We find the multi-teacher co-finetuning framework is very important.
This is because the hidden states of different teacher models can be in very different spaces, and jointly finetuning multiple teachers with shared pooling and prediction layers can align their output hidden spaces for better collaborative student teaching.
In addition, the distillation loss weighting method is also useful.
This is because the predictions of different teachers on the same sample may have different correctness, and focusing on the more reliable predictions is helpful for distilling accurate student models.

\begin{figure}[!t]
  \centering
  \subfigure[SST-2 and RTE datasets.]{
    \includegraphics[width=0.445\textwidth]{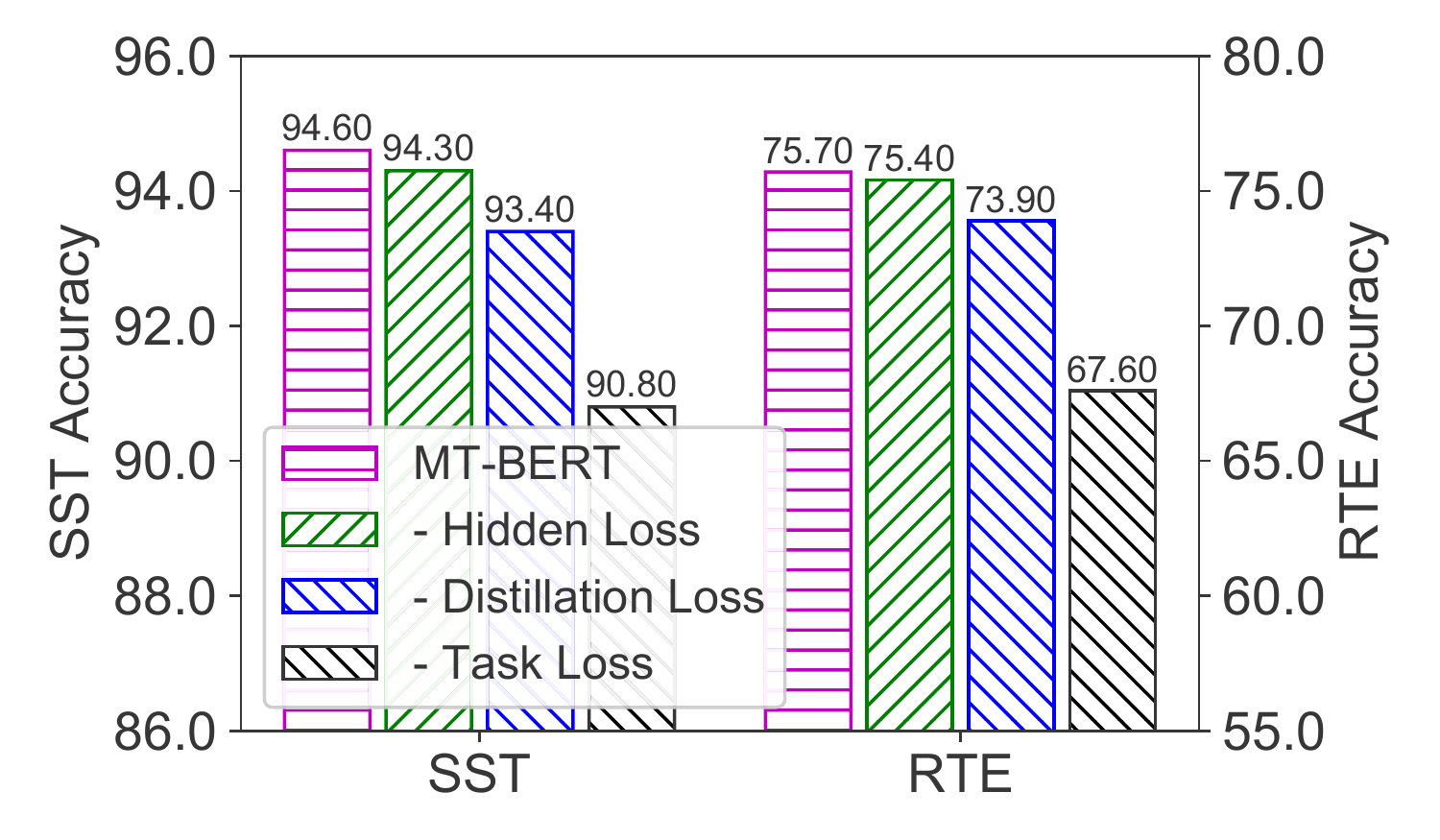}
  \label{fig.abl1}
  }
   \subfigure[MIND dataset.]{
      \includegraphics[width=0.445\textwidth]{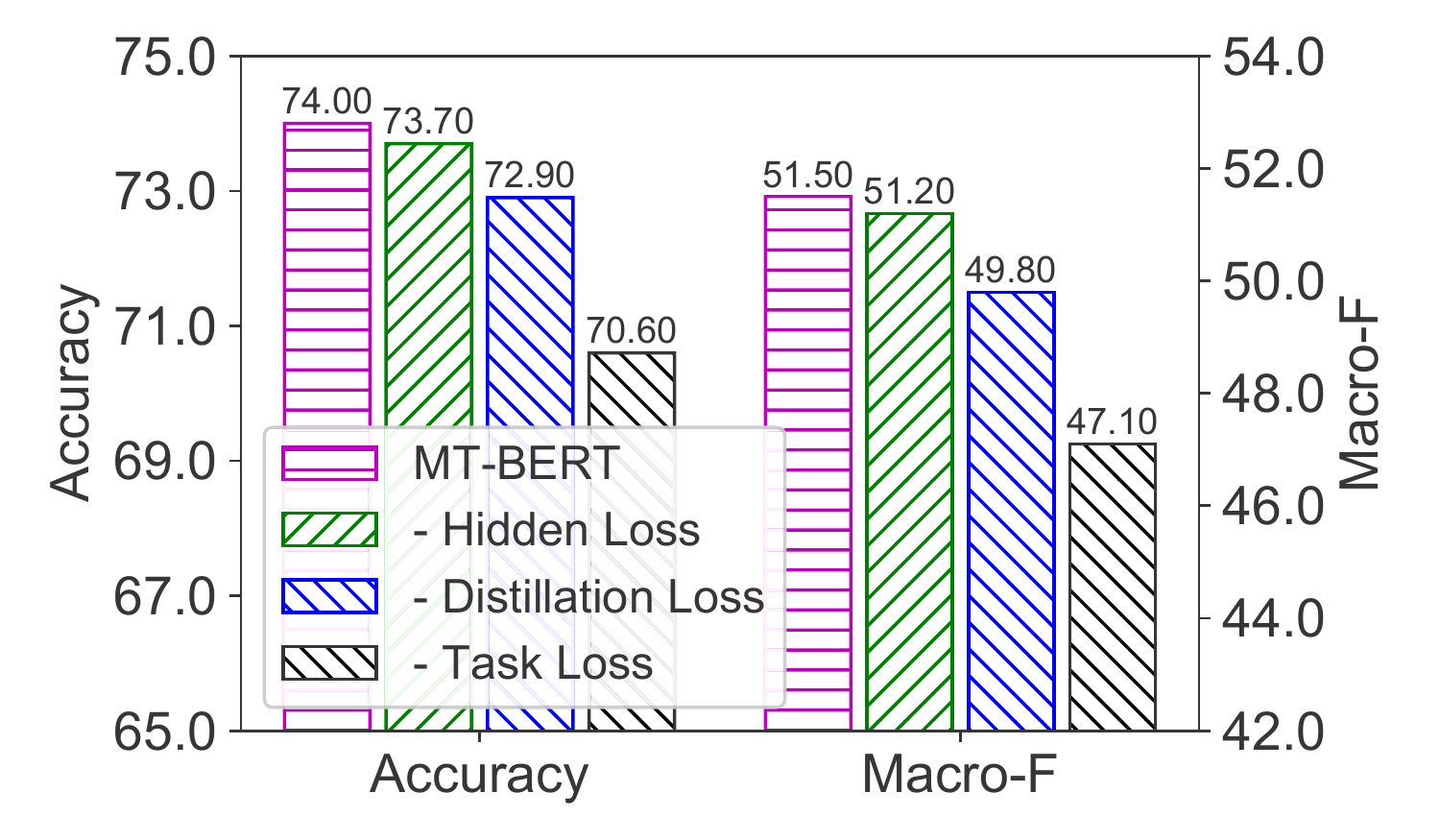}
  \label{fig.abl2}
  }
  \caption{Effectiveness of different loss functions.}\label{fig.abl}
\end{figure}

We also verify the effectiveness of different loss functions in MT-BERT, which is shown in Fig.~\ref{fig.abl}.
We find the task loss is very important.
It is because in our experiments the corpus for task-specific distillation are not large and the direct supervision from task labels is useful.
In addition, the distillation loss is also important.
It indicates that transferring the knowledge in soft labels plays a critical role in knowledge distillation.
Moreover, the hidden loss is also helpful.
It shows that hidden states of different teacher models can provide useful knowledge for student model learning.

\section{Conclusion}\label{sec:Conclusion}

In this paper, we propose a multi-teacher knowledge distillation method named MT-BERT for pre-trained language model compression, which can learn small but strong student model from multiple teacher PLMs in a collaborative way.
We propose a multi-teacher co-finetuning framework to align the output hidden states of multiple teacher models for better collaborative student teaching.
In addition, we design a multi-teacher hidden loss and a multi-teacher distillation loss to transfer the useful knowledge in both hidden states and prediction of multiple teacher models to student model.
The extensive experiments on three benchmark datasets show that MT-BERT can effectively improve the performance of pre-trained language model compression, and can outperform many single-teacher knowledge distillation methods.

\section*{Acknowledgments}
This work was supported by the National Natural Science Foundation of China under Grant numbers U1936208, U1936216, U1836204, and U1705261.
We thank Xing Xie, Tao Qi, Ruixuan Liu and Tao Di for their great comments and suggestions which are important for improving this work.

\bibliographystyle{acl_natbib}
\bibliography{acl2021}

\begin{thebibliography}{26}
\expandafter\ifx\csname natexlab\endcsname\relax\def\natexlab#1{#1}\fi

\bibitem[{Bao et~al.(2020)Bao, Dong, Wei, Wang, Yang, Liu, Wang, Gao, Piao,
  Zhou et~al.}]{bao2020unilmv2}
Hangbo Bao, Li~Dong, Furu Wei, Wenhui Wang, Nan Yang, Xiaodong Liu, Yu~Wang,
  Jianfeng Gao, Songhao Piao, Ming Zhou, et~al. 2020.
\newblock Unilmv2: Pseudo-masked language models for unified language model
  pre-training.
\newblock In \emph{ICML}, pages 642--652. PMLR.

\bibitem[{Bengio and LeCun(2015)}]{kingma2014adam}
Yoshua Bengio and Yann LeCun. 2015.
\newblock Adam: {A} method for stochastic optimization.
\newblock In \emph{ICLR}.

\bibitem[{Bentivogli et~al.(2009)Bentivogli, Clark, Dagan, and
  Giampiccolo}]{bentivogli2009fifth}
Luisa Bentivogli, Peter Clark, Ido Dagan, and Danilo Giampiccolo. 2009.
\newblock The fifth pascal recognizing textual entailment challenge.
\newblock In \emph{TAC}.

\bibitem[{Bhardwaj et~al.(2020)Bhardwaj, Majumder, and
  Poria}]{bhardwaj2020bias}
Rishabh Bhardwaj, Navonil Majumder, and Soujanya Poria. 2020.
\newblock Investigating gender bias in {BERT}.
\newblock \emph{arXiv preprint arXiv:2009.05021}.

\bibitem[{Devlin et~al.(2019)Devlin, Chang, Lee, and
  Toutanova}]{devlin2019bert}
Jacob Devlin, Ming-Wei Chang, Kenton Lee, and Kristina Toutanova. 2019.
\newblock Bert: Pre-training of deep bidirectional transformers for language
  understanding.
\newblock In \emph{NAACL-HLT}, pages 4171--4186.

\bibitem[{Dong et~al.(2019)Dong, Yang, Wang, Wei, Liu, Wang, Gao, Zhou, and
  Hon}]{dong2019unilm}
Li~Dong, Nan Yang, Wenhui Wang, Furu Wei, Xiaodong Liu, Yu~Wang, Jianfeng Gao,
  Ming Zhou, and Hsiao{-}Wuen Hon. 2019.
\newblock Unified language model pre-training for natural language
  understanding and generation.
\newblock In \emph{NeurIPS}, pages 13042--13054.

\bibitem[{Fukuda et~al.(2017)Fukuda, Suzuki, Kurata, Thomas, Cui, and
  Ramabhadran}]{fukuda2017efficient}
Takashi Fukuda, Masayuki Suzuki, Gakuto Kurata, Samuel Thomas, Jia Cui, and
  Bhuvana Ramabhadran. 2017.
\newblock Efficient knowledge distillation from an ensemble of teachers.
\newblock In \emph{Interspeech}, pages 3697--3701.

\bibitem[{Gou et~al.(2020)Gou, Yu, Maybank, and Tao}]{gou2020knowledge}
Jianping Gou, Baosheng Yu, Stephen~John Maybank, and Dacheng Tao. 2020.
\newblock Knowledge distillation: A survey.
\newblock \emph{arXiv preprint arXiv:2006.05525}.

\bibitem[{Jiao et~al.(2020)Jiao, Yin, Shang, Jiang, Chen, Li, Wang, and
  Liu}]{jiao2020tinybert}
Xiaoqi Jiao, Yichun Yin, Lifeng Shang, Xin Jiang, Xiao Chen, Linlin Li, Fang
  Wang, and Qun Liu. 2020.
\newblock Tinybert: Distilling {BERT} for natural language understanding.
\newblock In \emph{EMNLP Findings}, pages 4163--4174.

\bibitem[{Liu et~al.(2019)Liu, Ott, Goyal, Du, Joshi, Chen, Levy, Lewis,
  Zettlemoyer, and Stoyanov}]{liu2019roberta}
Yinhan Liu, Myle Ott, Naman Goyal, Jingfei Du, Mandar Joshi, Danqi Chen, Omer
  Levy, Mike Lewis, Luke Zettlemoyer, and Veselin Stoyanov. 2019.
\newblock Roberta: A robustly optimized bert pretraining approach.
\newblock \emph{arXiv preprint arXiv:1907.11692}.

\bibitem[{Liu et~al.(2020)Liu, Zhang, and Wang}]{liu2020adaptive}
Yuang Liu, Wei Zhang, and Jun Wang. 2020.
\newblock Adaptive multi-teacher multi-level knowledge distillation.
\newblock \emph{Neurocomputing}, 415:106--113.

\bibitem[{Lu et~al.(2020)Lu, Jiao, and Zhang}]{lu2020twinbert}
Wenhao Lu, Jian Jiao, and Ruofei Zhang. 2020.
\newblock Twinbert: Distilling knowledge to twin-structured compressed bert
  models for large-scale retrieval.
\newblock In \emph{CIKM}, pages 2645--2652.

\bibitem[{Qiu et~al.(2020)Qiu, Sun, Xu, Shao, Dai, and Huang}]{qiu2020pre}
Xipeng Qiu, Tianxiang Sun, Yige Xu, Yunfan Shao, Ning Dai, and Xuanjing Huang.
  2020.
\newblock Pre-trained models for natural language processing: A survey.
\newblock \emph{Science China Technological Sciences}, pages 1--26.

\bibitem[{Sanh et~al.(2019)Sanh, Debut, Chaumond, and
  Wolf}]{sanh2019distilbert}
Victor Sanh, Lysandre Debut, Julien Chaumond, and Thomas Wolf. 2019.
\newblock Distilbert, a distilled version of bert: smaller, faster, cheaper and
  lighter.
\newblock \emph{arXiv preprint arXiv:1910.01108}.

\bibitem[{Socher et~al.(2013)Socher, Perelygin, Wu, Chuang, Manning, Ng, and
  Potts}]{socher2013recursive}
Richard Socher, Alex Perelygin, Jean Wu, Jason Chuang, Christopher~D Manning,
  Andrew~Y Ng, and Christopher Potts. 2013.
\newblock Recursive deep models for semantic compositionality over a sentiment
  treebank.
\newblock In \emph{EMNLP}, pages 1631--1642.

\bibitem[{Sun et~al.(2019)Sun, Cheng, Gan, and Liu}]{sun2019patient}
Siqi Sun, Yu~Cheng, Zhe Gan, and Jingjing Liu. 2019.
\newblock Patient knowledge distillation for bert model compression.
\newblock In \emph{EMNLP-IJCNLP}, pages 4314--4323.

\bibitem[{Tang et~al.(2019)Tang, Lu, Liu, Mou, Vechtomova, and
  Lin}]{tang2019distilling}
Raphael Tang, Yao Lu, Linqing Liu, Lili Mou, Olga Vechtomova, and Jimmy Lin.
  2019.
\newblock Distilling task-specific knowledge from bert into simple neural
  networks.
\newblock \emph{arXiv preprint arXiv:1903.12136}.

\bibitem[{Wang et~al.(2020)Wang, Wei, Dong, Bao, Yang, and
  Zhou}]{wang2020minilm}
Wenhui Wang, Furu Wei, Li~Dong, Hangbo Bao, Nan Yang, and Ming Zhou. 2020.
\newblock Minilm: Deep self-attention distillation for task-agnostic
  compression of pre-trained transformers.
\newblock In \emph{NeurIPS}.

\bibitem[{Wu et~al.(2020{\natexlab{a}})Wu, Wu, Qi, Cui, and
  Huang}]{wu2020attentive}
Chuhan Wu, Fangzhao Wu, Tao Qi, Xiaohui Cui, and Yongfeng Huang.
  2020{\natexlab{a}}.
\newblock Attentive pooling with learnable norms for text representation.
\newblock In \emph{ACL}, pages 2961--2970.

\bibitem[{Wu et~al.(2020{\natexlab{b}})Wu, Wu, Qi, and Huang}]{wu2020improving}
Chuhan Wu, Fangzhao Wu, Tao Qi, and Yongfeng Huang. 2020{\natexlab{b}}.
\newblock Improving attention mechanism with query-value interaction.
\newblock \emph{arXiv preprint arXiv:2010.03766}.

\bibitem[{Wu et~al.(2021)Wu, Wu, Yu, Qi, Huang, and Liu}]{wu2021newsbert}
Chuhan Wu, Fangzhao Wu, Yang Yu, Tao Qi, Yongfeng Huang, and Qi~Liu. 2021.
\newblock Newsbert: Distilling pre-trained language model for intelligent news
  application.
\newblock \emph{arXiv preprint arXiv:2102.04887}.

\bibitem[{Wu et~al.(2020{\natexlab{c}})Wu, Qiao, Chen, Wu, Qi, Lian, Liu, Xie,
  Gao, Wu et~al.}]{wu2020mind}
Fangzhao Wu, Ying Qiao, Jiun-Hung Chen, Chuhan Wu, Tao Qi, Jianxun Lian,
  Danyang Liu, Xing Xie, Jianfeng Gao, Winnie Wu, et~al. 2020{\natexlab{c}}.
\newblock Mind: A large-scale dataset for news recommendation.
\newblock In \emph{ACL}, pages 3597--3606.

\bibitem[{Yang et~al.(2019)Yang, Dai, Yang, Carbonell, Salakhutdinov, and
  Le}]{yang2019xlnet}
Zhilin Yang, Zihang Dai, Yiming Yang, Jaime Carbonell, Russ~R Salakhutdinov,
  and Quoc~V Le. 2019.
\newblock Xlnet: Generalized autoregressive pretraining for language
  understanding.
\newblock In \emph{NeurIPS}, pages 5753--5763.

\bibitem[{Yang et~al.(2016)Yang, Yang, Dyer, He, Smola, and
  Hovy}]{yang2016hierarchical}
Zichao Yang, Diyi Yang, Chris Dyer, Xiaodong He, Alex Smola, and Eduard Hovy.
  2016.
\newblock Hierarchical attention networks for document classification.
\newblock In \emph{NAACL-HLT}, pages 1480--1489.

\bibitem[{You et~al.(2017)You, Xu, Xu, and Tao}]{you2017learning}
Shan You, Chang Xu, Chao Xu, and Dacheng Tao. 2017.
\newblock Learning from multiple teacher networks.
\newblock In \emph{KDD}, pages 1285--1294.

\bibitem[{Yuan et~al.(2020)Yuan, Shou, Pei, Lin, Gong, Fu, and
  Jiang}]{yuan2020reinforced}
Fei Yuan, Linjun Shou, Jian Pei, Wutao Lin, Ming Gong, Yan Fu, and Daxin Jiang.
  2020.
\newblock Reinforced multi-teacher selection for knowledge distillation.
\newblock \emph{arXiv preprint arXiv:2012.06048}.

\end{thebibliography}


\end{document}